\documentclass[12pt,a4paper,twocolumn]{narms}

\usepackage{epsfig}
\usepackage{timesmt}
\usepackage{siunitx}
\usepackage{xcolor}

\usepackage{caption} 
\usepackage{subcaption} 
\usepackage{import} 

\usepackage{multirow}	
\usepackage{booktabs}

\usepackage{newtxtext,newtxmath} 
\usepackage{amsmath} 
\usepackage{hyperref} 

\usepackage{00_Template/chicaco}

\DeclareAcronym{2D}{
	short   = 2D,
	long    = Zweidimensional,
	}
	
\DeclareAcronym{3D}{
	short   = 3D,
	long    = Dreidimensional,
	}

\DeclareAcronym{BIM}{
	short   = BIM,
	long    = Building Information Modeling,
	}
	
\DeclareAcronym{BIM model}{
	short   = BIM model,
	long    = building information model,
	}
	
\DeclareAcronym{URDF}{
	short   = URDF,
	long    = Universal Robot Description Format,
	}

\DeclareAcronym{SDF}{
	short   = SDF,
	long    = Simulation Definition Format,
	}
	
\DeclareAcronym{SVG}{
	short   = SVG,
	long    = Scalable Vector Graphics,
	}
	
\DeclareAcronym{BIRS}{
	short   = BIRS,
	long    = Building Information Robotic System,
	}

\DeclareAcronym{Scan-vs-BIM}{
	short   = Scan-vs-BIM,
	long    = comparison between a point cloud and a \ac{BIM} model,
	}

\DeclareAcronym{GIS}{
	short   = GIS,
	long    = Geographic Information System,
	}
	
\DeclareAcronym{GeoLab}{
    short = GeoLab,
    long  = Geodätisches Labor,
    long-genitive-form = Geodätischen Labors,
    long-dative-form = Geodätischen Labor,
    }
    
\DeclareAcronym{BGU}{
	short   = BGU,
	long    = \DAfaculty{},
	}
    
\DeclareAcronym{EMM}{
	short   = EMM,
	long    = Environment Mapping Module,
	}
	
\DeclareAcronym{SLAM}{
	short   = SLAM,
	long    = Simultaneous Localization and Mapping,
	}
	
\DeclareAcronym{V-SLAM}{
	short   = V-SLAM,
	long    = Visual SLAM,
	}
	
\DeclareAcronym{UGV}{
	short   = UGV,
	long    = Unmanned Ground Vehicle ,
	}	
	
\DeclareAcronym{UAV}{
	short   = UAV,
	long    = Unmanned Aerial Vehicle,
	}	
		
 \DeclareAcronym{UV}{
	short   = UV,
	long    = Unmanned Vehicle,
	}	
	
\DeclareAcronym{UVs}{
	short   = UVs,
	long    = Unmanned Vehicles,
	}
	
\DeclareAcronym{LiDAR}{
	short   = LiDAR,
	long    = Light Detection and Ranging,
	}

\DeclareAcronym{SBAS}{
	short   = SBAS,
	long    = Satellite Based Augmentation Systems,
	}	
	
\DeclareAcronym{IMU}{
	short   = IMU,
	long    = Inertial Measurement Units,
	}

\DeclareAcronym{GNSS}{
	short   = GNSS,
	long    = Global Navigation Satellite System,
	}		

\DeclareAcronym{GPS}{
	short   = GPS,
	long    = Global Positioning System,
	}	
	
\DeclareAcronym{D-GPS}{
	short   = D-GPS,
	long    = Differential \ac{GPS},
	}	
	
\DeclareAcronym{MAP}{
	short   = MAP,
	long    = Maximum A Posteriori,
	}		
	
\DeclareAcronym{EKF}{
	short   = EKF,
	long    = Extended Kalman Filter,
	}	
		
\DeclareAcronym{BA}{
	short   = BA,
	long    = Bundle-Adjustment,
	}	
		
\DeclareAcronym{DNN}{
	short   = DNN,
	long    = Deep Neural Network,
	}	
	
\DeclareAcronym{GNN}{
	short   = GNN,
	long    = Graph neural networks,
	}	
 
\DeclareAcronym{DL}{
	short   = DL,
	long    = Deep Learning,
	}	
	
\DeclareAcronym{UWB}{
	short   = UWB,
	long    = Ultra Wide Band,
	}	
	
\DeclareAcronym{TLS}{
	short   = TLS,
	long    = Terrestrial Laser Scanner,
	}
	
\DeclareAcronym{MLS}{
	short   = MLS,
	long    = Mobile Laser Scanner,
	}

\DeclareAcronym{SAR}{
	short   = SAR,
	long    = Search and Rescue,
	}	
	
\DeclareAcronym{SfM}{
	short   = SfM,
	long    = Structure from Motion,
	}

\DeclareAcronym{MVS}{
	short   = MVS,
	long    = Multi-View Stereo,
	}

\DeclareAcronym{KPIs}{
	short   = KPIs,
	long    = Key Performance Indicators,
	}

\DeclareAcronym{MVE}{
	short   = MVE,
	long    = Multiview Environment,
	}	
	
\DeclareAcronym{RTPS}{
	short   = RTPS,
	long    = Real Time Positioning System,
	}

\DeclareAcronym{ICP}{
	short   = ICP,
	long    = Iterative Closest Point,
	}

\DeclareAcronym{GICP}{
	short   = GICP,
	long    = Generalized ICP,
	}	
	
\DeclareAcronym{AMCL}{
	short   = AMCL,
	long    = Adaptive Monte Carlo Localization,
	}	
	
\DeclareAcronym{GMCL}{
	short   = GMCL,
	long    = General Monte Carlo Localization,
	}	
	
\DeclareAcronym{SER}{
	short   = SER,
	long    = Similar Energy Region,
	}	
		
\DeclareAcronym{PF}{
	short   = PF,
	long    = Particle Filter,
	}	
	
\DeclareAcronym{RANSAC}{
	short   = RANSAC,
	long    = Random Sample Consensus,
	}	
	
\DeclareAcronym{ROS}{
	short   = ROS,
	long    = Robot Operating System,
	}	

\DeclareAcronym{DoF}{
	short   = DoF,
	long    = Degrees of Freedom,
	}
	
\DeclareAcronym{MAV}{
	short   = MAV,
	long    = Micro Aerial Vehicle,
	}	
	
\DeclareAcronym{VP}{
	short   = VP,
	long    = Vanishing Points,
	}	
	
\DeclareAcronym{VL}{
	short   = VL,
	long    = Vanishing Lines,
	}	
	
\DeclareAcronym{VR}{
	short   = VR,
	long    = Virtual Reality,
	}	
	
\DeclareAcronym{AR}{
	short   = AR,
	long    = Augmented Reality,
	}	
	
\DeclareAcronym{MR}{
	short   = MR,
	long    = Mixed Reality,
	}	
	
\DeclareAcronym{LoD}{
	short   = LoD,
	long    = Level of Detail,
	}	
	
\DeclareAcronym{IFC}{
	short   = IFC,
	long    = Industry Foundation Classes,
	}	
	
\DeclareAcronym{CPS}{
	short   = CPS,
	long    = Cyber-Physical Systems,
	}		
	
\DeclareAcronym{LOAM}{
	short   = LOAM,
	long    = LiDAR Odometry and Mapping,
	}	
	
\DeclareAcronym{A-LOAM}{
  short = A-LOAM,
  long = Advanced implementation of LOAM
  }
  
\DeclareAcronym{F-LOAM}{
  short = F-LOAM,
  long = Fast LiDAR Odometry And Mapping
  }

\DeclareAcronym{IFR}{
  short = IFR,
  long = International Federation of Robotics
  }

\DeclareAcronym{RGB-D}{
  short = RGB-D,
  long = Red-Green-Blue-Depth
  }
   
\DeclareAcronym{OGM}{
  short = OGM,
  long = Occupancy Grid Map
  }
  
\DeclareAcronym{KLD}{
  short = KLD,
  long = Kullback-Leibler distance
  }
    
 \DeclareAcronym{MCL}{
  short = MCL,
  long = Monte Carlo Localization
  }
    
 \DeclareAcronym{GBL}{
  short = GBL,
  long = Graph-based Localization
  }     
  
\DeclareAcronym{RViz}{
  short = RViz,
  long = ROS visualization
  }

\DeclareAcronym{APE}{
  short = APE,
  long = Absolute Pose Error
  }  

\DeclareAcronym{RE}{
  short = RE,
  long = Rotational Error 
  } 
  
  \DeclareAcronym{RMSE}{
  short = RMSE,
  long =  Root Mean Square Error 
  }

\begin{document}

\title{Occupancy Grid Map to Pose Graph-based Map: Robust BIM-based 2D-LiDAR Localization for Lifelong Indoor Navigation in Changing and Dynamic Environments} 

\author{{M.A. Vega Torres 
\& A. Braun 
\& A. Borrmann} \\
{\aff{Chair of Computational Modeling and Simulation}} \\
{\aff{Technical University of Munich, Munich, Germany}
}}

\date{}
\abstract{
Several studies rely on the de facto standard \ac{AMCL} method to localize a robot in an \ac{OGM} extracted from a \ac{BIM model}. 
However, most of these studies assume that the \ac{BIM model} precisely represents the real world, which is rarely true.
Discrepancies between the reference \ac{BIM model} and the real world (Scan-BIM deviations) are not only due to the presence of furniture or clutter but also due to the usual as-planned and as-built deviations that exist with any model created in the design phase.
These Scan-BIM deviations may affect the accuracy of \acs{AMCL} drastically.
This paper proposes an open-source method to generate appropriate Pose Graph-based maps from BIM models for robust 2D-LiDAR localization in changing and dynamic environments.
First, 2D \acs{OGM}s are automatically generated from complex \acs{BIM model}s. These \acs{OGM}s only represent structural elements allowing indoor autonomous robot navigation.
Then, an efficient technique converts these 2D \acs{OGM}s into Pose Graph-based maps enabling a more accurate robot pose tracking. 
Finally, we leverage the different map representations for accurate, robust localization with a combination of state-of-the-art algorithms.
Moreover, we provide a quantitative comparison of various state-of-the-art localization algorithms in three simulated scenarios with varying levels of Scan-BIM deviations and dynamic agents. 
More precisely, we compare two \ac{PF} algorithms: \acs{AMCL} and \ac{GMCL}; and two \ac{GBL} methods: Google's Cartographer and SLAM Toolbox, solving the global localization and pose tracking problems.
We found that in a real office environment (under medium level of Scan-BIM deviations) the translational RMSE of \acs{AMCL} increases by a factor of four (from \SI{8,5}{\centi\metre} in the empty environment to \SI{33,7}{\centi\metre} in the real one). 
On the contrary, pose Graph-based algorithms demonstrate their superiority in contrast to particle filter (PF) algorithms, achieving an RMSE of \SI{7,2}{\centi\metre}, even in the real environment.
The numerous experiments demonstrate that the proposed method contributes to a robust localization with an as-designed \acs{BIM model} or a sparse \acs{OGM} in changing and dynamic environments, outperforming the conventional \acs{AMCL} in accuracy and robustness.
}

\maketitle

\section{INTRODUCTION}
An accurate localization system is crucial for successful autonomous mobile robot deployment in indoor GPS-denied environments.

The indoor localization problem has been approached by applying several techniques.
While some of them rely on the known position of landmarks, such as AprilTags or textual cues, others depend on sensors that have to be installed strategically on the building, such as Beacons or WiFi access points.
However, in most cases, the exact location of specific landmarks is not known in advance.
On top of that, having additional sensors increases the cost of the navigation stack.

A \acs{BIM model}, which is available for most of the current architecture, can be used as a reference map for LiDAR localization. 
Moreover, the additional semantic information of the model can be exploited to create advanced automated robotic tasks, like object inspection \shortcite{Kim.2022} or painting \shortcite{Kim.2021}, which at the same time depend on an accurate localization system.

The main issue of using a \ac{BIM model} or a floor-plan as a reference map for 2D-LiDAR localization is the presence of Scan-BIM deviations. 
These deviations can be caused by furniture or clutter not present in the model, as-planned and as-built deviations, and dynamic or ``quasi-static'' changes in the environment.
In an effort to address this challenge, we contribute with a system that creates \acs{OGM}s from \ac{BIM model}s and allows their automatic transformation in pose graph-based maps. 
These maps are leveraged for quick, memory efficient, and accurate localization in indoor GPS-denied environments, enabling safer autonomous navigation.

More specifically, the following are our contributions:

\begin{itemize}
    \item A method to extract \ac{OGM}s out of complex multi-story BIM models to allow path planning and autonomous navigation of robots in indoor GPS-denied environments.
    
    \item An efficient open-source method \footnote{Available at: \url{https://github.com/MigVega/Ogm2Pgbm}} to convert these 2D \ac{OGM}s into pose graph-based maps for accurate 2D-LiDAR localization and navigation.
    
    \item An extensive quantitative comparison of various state-of-the-art 2D LiDAR localization algorithms in three carefully designed simulated scenarios with different levels of Scan-BIM deviations and with and without dynamic agents. 
  
\end{itemize}

The remainder of this paper is organized as follows.
Section \ref{chap:theory} introduces the problem formulation of LiDAR localization as well as the main principles behind the particle filter-based and the graph-based localization strategies.
Section \ref{Chap:related_work} describes previous work done on BIM-based LiDAR localization. 
Section \ref{chap:methodology} introduces our method to generate \ac{OGM}s from \ac{BIM model}s and pose graph-based maps from \ac{OGM}s, as well as the proposed employment of these maps for robust localization.
Section \ref{chap:experiments} presents the experimental settings, followed by the results and analysis in section \ref{chap:results}. 
Finally, section \ref{chap:conclusions} concludes our work.
\section{THEORETICAL BACKGROUND}
\label{chap:theory}

Before presenting current state-of-the-art methods, a brief introduction to the theoretical basis behind the two main types of localization algorithms used in this research is provided.

\subsection{Localization problem}
In this paper, we address the robot pose tracking and global localization problems, i.e., with and without approximated initial pose respectively, given a 3D BIM model as a prior map which omits considerable information about the real environment and assuming that the robot employs a 2D-LiDAR sensor.

In the 2D problem, the pose of the robot at time $t$ is defined as position and orientation $\boldsymbol{x}_{t}=[x, y, \theta]^{\top}$ in the coordinate system of the map.

We aim to estimate the most likely robot's pose $\boldsymbol{x}_{t}^{*}$ given the measurements $\boldsymbol{z}_{t}$ and the map $\boldsymbol{m}$.
Formally, the goal is to compute:

\begin{align}\label{eq:localization}
\boldsymbol{x}_{t}^{*}=\underset{\boldsymbol{x}}{\arg \max } p\left(\boldsymbol{x}_{t} \mid \boldsymbol{z}_{t}, \boldsymbol{m}\right)
\end{align}

Two widely used methods that aim to calculate this estimate are the \acf{PF} and the \acf{GBL} algorithms.

\subsection{Particle Filter algorithms}

\ac{PF} algorithms, also called \ac{MCL} methods, are probabilistic approaches that represent the pose estimate with a set of normalized weighted particles. 
Each particle $\boldsymbol{s}_{t}^{i}=\left\langle\boldsymbol{x}_{t}^{i}, \omega_{t}^{i}\right\rangle$
consist of a pose $\boldsymbol{x}_{t}^{i}$ and a weight $\omega_{t}^{i}$.

Initially, a set $\mathcal{M}$ of particles is sampled from a Gaussian distribution around the possible locations of the robot.
Subsequently, three steps are repeated iteratively in the algorithm: motion update, importance weighting, and particle resampling.
For a more detailed explanation of every step, the reader is referred to \shortciteN{Thrun.2005}.
In our comparison we implemented \ac{AMCL} \shortcite{Pfaff.2006} and \ac{GMCL} \shortcite{AlshikhKhalil.2021} to be tested under different levels of Scan-BIM deviations.

\subsection{Graph-based algorithms}
Graph-based, also called optimization-based localization methods use previously acquired pose-graph data for pose estimation. 
This pose-graph contains landmarks of the environment (which can be represented as submaps) associated with nodes (which are the poses from where the landmarks were observed). 
Additionally, the nodes are bound to each other with spatial constraints. 
In a sliding window manner, the method not only considers the most recent measurement but a set of them to compute the current pose.

Under the assumption that the measurements are normally distributed and i.i.d., it is possible to represent eq. \ref{eq:localization} as a weighted least squares problem.  This problem is commonly solved iteratively using the Levenberg-Marquart algorithm.

In this paper we compare Cartographer \shortcite{Hess.2016b} and SLAM Toolbox \shortcite{Macenski.2021} as \ac{GBL} algorithms.

While particle filter algorithms are easier to implement and can represent non-Gaussian distributions, graph-based localization algorithms, besides being deterministic, can handle delayed measurements and maintain a recent history of poses. A more exhaustive qualitative comparison is given by \shortciteN{Wilbers.2019}. 

\section{RELATED RESEARCH}
\label{Chap:related_work}

A \ac{BIM model} with 3D geometric information can be used as a prior map to accurately localize robots in indoor GPS-denied environments and allow autonomous navigation.
This section will overview state-of-the-art methods, which used prior building information, i.e., \ac{BIM model}s or floor plans, to find the correct robot position and orientation.

\shortciteN{Follini.2020.bim} show that the transformation matrix between the reference system of the robot and the map extracted from the BIM model can be retrieved by applying the standard \ac{AMCL} algorithm.

The same algorithm was used by \shortciteN{Prieto.2020}, \shortciteN{Karimi.21.04.2021}, \shortciteN{Kim.2021}, and \shortciteN{Kim.2022} to localize a wheeled robot in an \ac{OGM} generated from the \ac{BIM model}.

The main difference between these methods relies on how they create the \ac{OGM} from the BIM model.

While \shortciteANP{Follini.2020.bim} took the vertices of elements that intersect a horizontal plane and used the Open CASCADE viewer to generate an \ac{OGM} in \textit{pgm} format with the corresponding resolution and map origin information, \shortciteANP{Prieto.2020} uses the geometry of the spaces in the \ac{IFC} file and the location and size of each one of the openings.

\shortciteN{Karimi.2020} developed a \ac{BIRS}, enabling the generation and semantic transfer of topological and metric maps from a BIM model to \ac{ROS}.  
The tool was further developed in \shortcite{Karimi.21.04.2021} with an optimal path planner, integrating critical components for construction assessment. 

\shortciteN{Kim.2021} implemented a method to convert an \ac{IFC} file into a \ac{ROS}-compliant \ac{SDF} world file suitable for robot task planning. They evaluated their approach for the purpose of indoor wall painting. 

Later, to incorporate dynamic objects and for the aim of door inspection, \shortciteN{Kim.2022} proposed a technique to convert an \ac{IFC} model into a \ac{URDF} building world. 
Once they have the \ac{URDF} model, they use the PgmMap creator \shortcite{Yang.2018} to create an \ac{OGM} out of it.

\shortciteN{Hendrikx.2021} proposed an approach that instead of using an \ac{OGM} uses a robot-specific world model representation extracted from an \ac{IFC} file for 2D-LiDAR localization. 

In their factor graph-based localization approach, the system queries semantic objects in its surroundings and creates data associations between them and the laser measurements. 

While they demonstrated that the method can track the pose of the robot, it was  not evaluated quantitatively.

Instead of using a BIM model, \shortciteN{Boniardi.2017} use a CAD-based architectural floor plan for 2D LiDAR-based localization. In their localization system, they implement \ac{GICP} for scan matching together with a pose graph \ac{SLAM} system. 

Later, they proposed an improved pipeline for long-term localization in dynamic environments \shortciteN{Boniardi.2019}.

\shortciteN{Zimmerman.23.03.2022} uses an \ac{OGM} obtained from a sliced \ac{TLS} point cloud a together with human-readable localized cues to assist localization. With their text detection-based localization technique, they can detect known room numbers and thus can robustly handle symmetric environments with structural changes.

While several approaches have emerged aiming to create \ac{OGM}s from BIM models, none of them deal with complex non-convex models with multiple stories and slanted floors.
Moreover, most of the proposed techniques are based on the strong assumption that the BIM model represents the actual current state of the building very precisely, ignoring the presence of possible Scan-BIM deviations due to clutter, furniture, as-planned vs. as-built differences, changes due to long-term operation, or the presence of dynamic agents. 
\section{METHODOLOGY}
\label{chap:methodology}


Our method can be divided in three main steps:
\textbf{Step 1:} Creation of an \ac{OGM} from an \ac{IFC} file employing IfcConvert and OpenCV. 
\textbf{Step 2:} Automatic generation of a Pose Graph-based map out of an \ac{OGM} with a combination of image processing, coverage path planner, and ray casting. 
\textbf{Step 3:} Robust localization using particle filter algorithm and graph-based localization system.

\begin{figure*}
\centerline{
\includegraphics[width=1.05\textwidth]{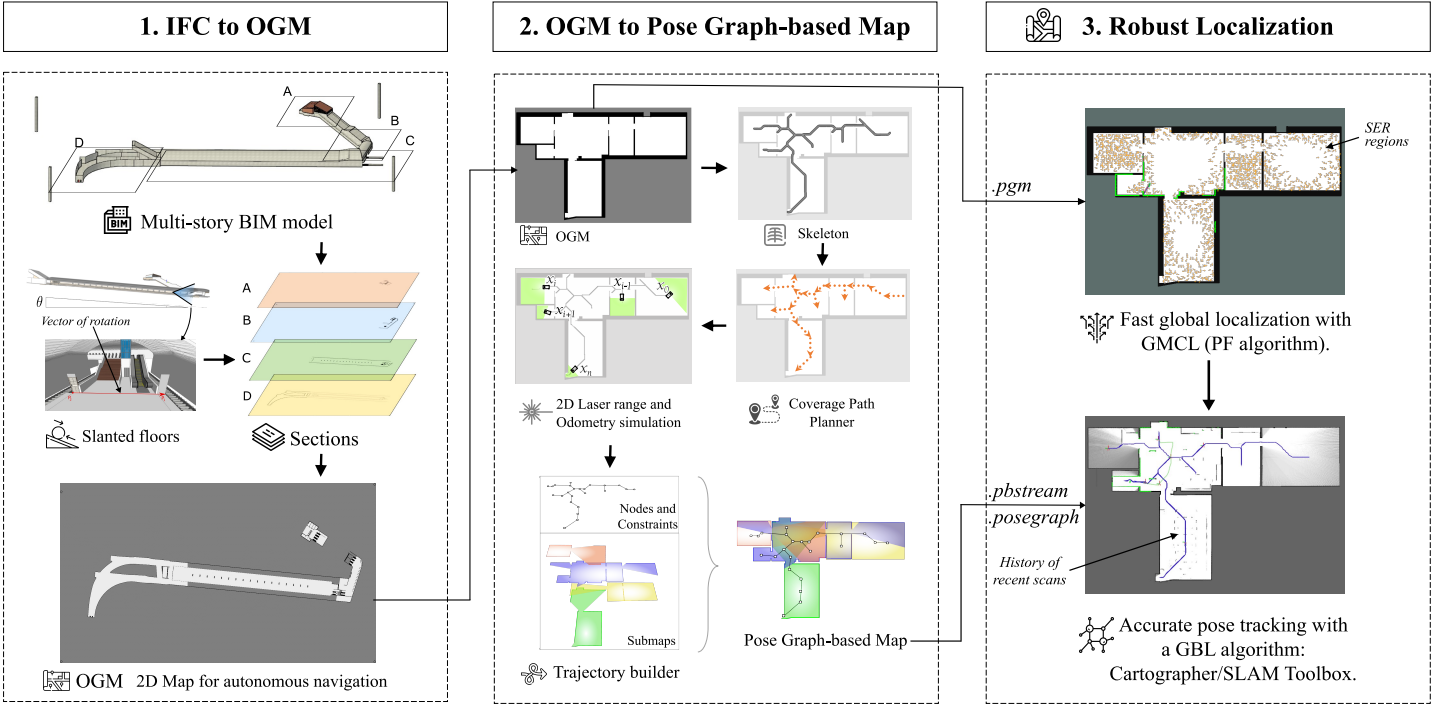}
}
\caption{Proposed IFC to Pose Graph-based map for robust 2D-LiDAR localization. In the first step, an OGM is created from Multi-story non-convex BIM models which can have slanted floors, this map is suitable for path planning and autonomous robot navigation. In the second step, a Pose Graph-based map is generated from the OGM. Finally, in the third step, these maps allow fast global localization and robust pose tracking in changing and dynamic environments.}
\label{fig:method}
\end{figure*}

\subsection{\ac{OGM} generation from an \ac{IFC}}
For the creation of suitable 2D \ac{OGM} for robot localization and navigation from a complex multi-story \ac{IFC} models the IfcConvert tool of IfcOpenSchell \shortcite{krijnen2015ifcopenshell} and image processing techniques are used.

IfcConvert allows the creation of a 2D map in \ac{SVG} format with the desired elements in the \ac{IFC} model that cross a plane at the desired height.

In our case, non-permanent entities such as spaces, windows, and doors are excluded from the resulting 2D \ac{OGM} by ignoring the corresponding entity names.
This exclusion is essential to filter only structural information about the building, enabling further autonomous navigation between the rooms that want to be explored. 
Besides having the permanent structures in the \ac{OGM}, and with the aim of global localization and posterior correct pose graph map generation, it is crucial to differentiate between outdoor (unknown) and indoor (navigable) spaces in the \ac{OGM}. 
This distinction can be automated creating a second \ac{OGM} with all the entities in the \ac{IFC} file (i.e., with doors, windows, and spaces). 

The final separation of outdoor (gray color), indoor (white) and obstacle (black) is done based on the contours in the \ac{SVG} image. 
OpenCV allows the processing of the contours depending on their hierarchy, i.e., depending if they are inside  (child contours) or outside another contour (parent contours).

The resulted file is finally converted to \textit{.pgm}, which together with its properties (the resolution and origin) in a \textit{.yaml} file can then be loaded in the robotic system as prior environment information, allowing robot localization, path planning, and autonomous navigation.

A similar procedure can be followed for multi-story level buildings. 
In the particular case of non-overlapping stories, the different \ac{OGM}s can be merged in a single one if the relative position between them is known. 
To maintain this spacial relationship, while obtaining the \ac{OGM}s, reference auxiliary elements with a height equal to the maximum building's height can be included in its surroundings. With these additional elements, all the \ac{OGM}s will have the same dimensions allowing its merging.

Creating 2D \ac{OGM}s with IfcConvert is relatively straightforward when the desired section is horizontal (parallel with the XY plane). 
However, if the model has a ramp or a slightly slanted floor, the model must be rotated before the occupancy map is generated. 

Favorably, IfcConvert also allows the rotation of the model at the desired angle given a quaternion calculated from the vector of rotation.

\subsection{OGM2PGBM: \ac{OGM} to Pose Graph-based map conversion }

The automatic generation of data suitable for \ac{GBL} methods from BIM models implies the simulation of sequential laser data in the entire navigable space in the model with the corresponding odometry data.

For this aim, the previously generated 2D \ac{OGM} are used.

Applying the skeleton method proposed by \shortciteN{lee1994building} enables the interconnection of all the rooms in a smooth trajectory. 

Subsequently, a Wavefront Coverage Path Planner \shortciteN{Zelinsky.1993} is applied over the navigable area inside a dilated version of the skeleton, allowing finding the waypoints over which the laser will be simulated. 

Then, using a ray casting algorithm and without a real-time simulation engine (such as Gazebo), laser sensor data and odometry are simulated following the waypoints found in the previous step. 
Finally, a trajectory builder merges these sensor data creating an accurate pose graph-based map, serialized as a \textit{.pbstream} file for Cartographer or as a \textit{.posegraph} file for SLAM Toolbox.
Graph optimization is not required since every scan's position is known accurately from the simulation. 

This pipeline allows the automatic efficient generation of pose graph-based maps (with submaps, nodes, and constraints) from a 2D \ac{OGM}. 
As our OGM2PGBM workflow does not require Gazebo for data simulation, it is faster and more portable than a Gazebo-based pipeline, allowing its execution on an isolated manner. 
Moreover, since the technique does not consider the complete 3D model but only a 2D \ac{OGM}, it is very efficient. 
In addition, it can be used from any given \ac{OGM}, which besides of been generated from a BIM model (with the method presented in the previous section), can be generated out of a floor plan or a previously scanned map.


\subsection{Robust Localization}

Once the different needed map representations (OGM and pose graph-based maps) are generated from a BIM model, they can be used for robust localization in changing environments.
We propose to take advantage of the Self-Adaptive \ac{PF} of \ac{GMCL} to spread particles only in the \ac{SER} regions and solve the global localization problem efficiently.
As it is shown later (in Section \ref{chap:results}), \ac{PF} algorithms being able of representing non-Gaussian distributions can solve the global localization faster than graph-based algorithms.
Once an estimated pose is found with a covariance smaller than 0.05, the nodes of \ac{GMCL} are stopped, and a \ac{GBL} algorithm can be started. 
For example, to track the pose of the robot accurately, Cartographer can be activated with the \textit{start\_traj} service at the time when \ac{GMCL} converges and using the \textit{.pbstream} map generated with the method proposed in Section 4.3.
Similarly, SLAM Toolbox can be started with an initial pose, however with a prior \textit{.posegraph} map.
\section{EXPERIMENTS}
\label{chap:experiments}

This section presents the evaluation scenarios designed to evaluate the various techniques and details of the implementation and evaluation.

\subsection{Evaluation Scenarios}
\begin{figure}[htb]
  \centering
    \begin{subfigure}[b]{0.15\textwidth}
        \includegraphics[width=\textwidth]{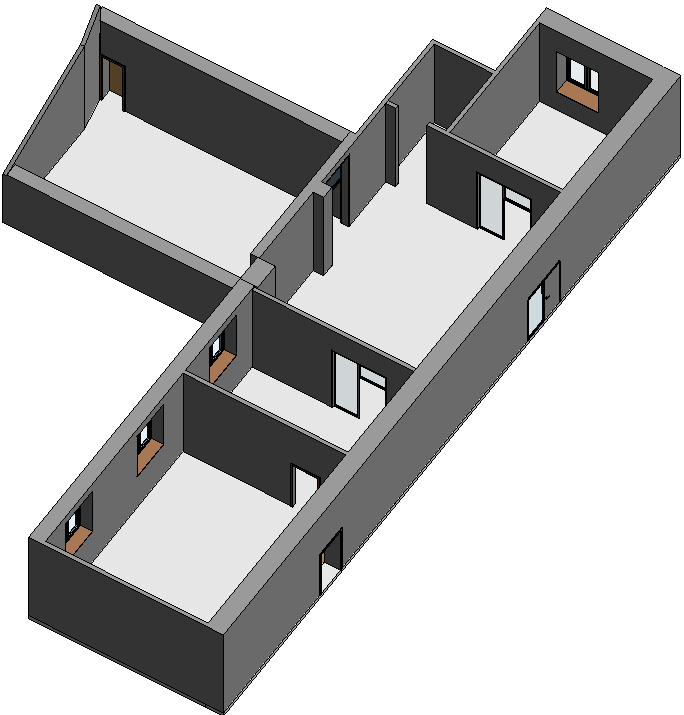}
        \caption{1} \label{fig:Scenario1}
    \end{subfigure}
    \begin{subfigure}[b]{0.15\textwidth}
        \includegraphics[width=\textwidth]{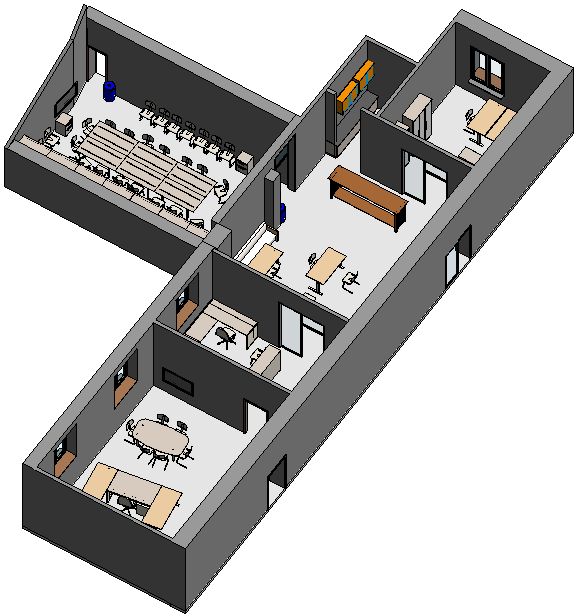}
        \caption{2} \label{fig:Scenario2}
    \end{subfigure}
    \begin{subfigure}[b]{0.15\textwidth}
        \includegraphics[width=\textwidth]{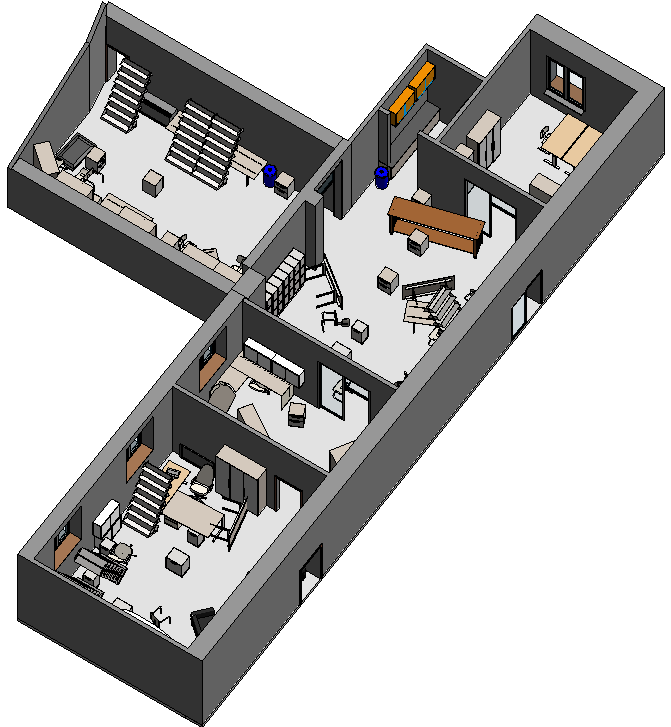}
        \caption{3} \label{fig:Scenario3}
    \end{subfigure}
\caption[Evaluation Scenarios]{Evaluation Scenarios. (a) Empty Room: represents a typical BIM Model, without furniture; (b) Reality: represents a standard office environment and is based on real-world \ac{TLS} data; (c) Disaster: is an environment after a simulated disaster with large Scan-BIM deviations.} \label{fig:Scenarios}
\end{figure}

As illustrated in Figure \ref{fig:Scenarios}, three different scenarios were conceived to evaluate the different methods. Each scenario increases the level of clutter present in the environment and, therefore, decreases the level of overlap that a perception sensor would have with permanent building objects (such as walls, columns, floors, and ceilings). The latter are the elements that are usually present in a BIM model.

\nointerlineskip

\begin{figure*}[hbt]
\centering
\subfloat[1-1]{\label{fig:Sequence1_1}{
	\includegraphics[width=0.14\textwidth]{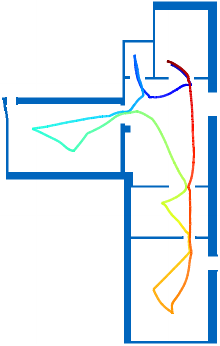}}}\hfill
\subfloat[1-2]{\label{fig:Sequence1_2}{
	\includegraphics[width=0.14\textwidth]{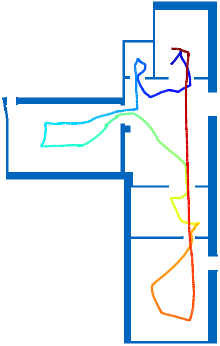}}}\hfill
\subfloat[2-1]{\label{fig:Sequence2_1}{
    \includegraphics[width=0.14\textwidth]{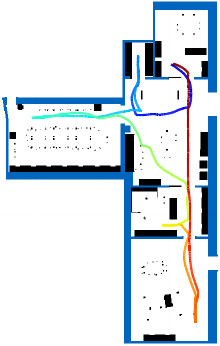}}}\hfill
\subfloat[2-2]{\label{fig:Sequence2_2}{
	\includegraphics[width=0.14\textwidth]{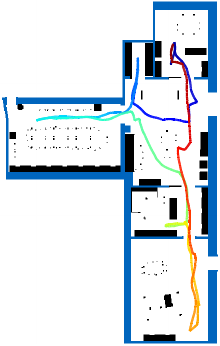}}}\hfill
\subfloat[3-1]{\label{fig:Sequence3_1}{
	\includegraphics[width=0.14\textwidth]{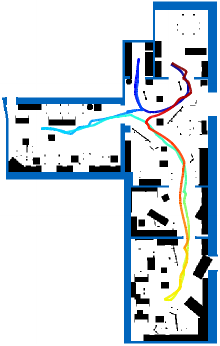}}}\hfill
\subfloat[3-2]{\label{fig:Sequence3_2}{
	\includegraphics[width=0.14\textwidth]{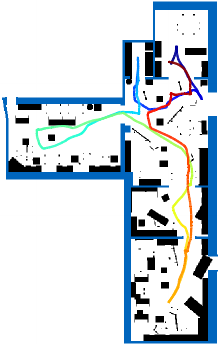}}}
\caption[Sequences]{Sequences of data with the respective \ac{OGM}s. (a) and (b) correspond to an empty environment (i.e., without furniture) with and without dynamic agents resp.; (c) and (d) similar but in an scenario with furniture as it is in the real world; (e) and (f) in the disaster environment. To better visualize the different levels of Scan-BIM deviations, the \ac{OGM} of the empty environment is presented over the other \ac{OGM}s in blue color. The change in color of the trajectory represents the initial and end position of the robot, with dark blue being the start and red the endpoint.}\label{fig:Sequences}
\end{figure*}

Additionally, to increase the simulation's realism level, we added animated walking human models (also called dynamic agents) moving in the environment. 
In scenarios 1 and 2, five humans walk from each room to the closest exit of that room. 
In the scenario Nr. 3 (``Disaster"), a total of six people move faster, trying to escape through the main door. 
Once the agents reach their goal, they start again, moving from their initial planned position in an infinite loop.

\subsection{Gazebo Simulation}

To simulate the experimental data, we use Gazebo. 
Once the \ac{IFC} model is converted to Collada format using IfcConvert, it can be imported in Gazebo.
While importing complex \ac{IFC} models in Gazebo is essential to ensure that every element has its own geometric representation. 
One way to avoid instantiating multiple objects from the same data is using the export capabilities of Blender. 

For trustworthy data simulation, we separate between collision and visual models. 
Since LiDAR sensors cannot perceive glass materials, windows and glass doors were removed in the collision models.

\subsection{Robot Simulation}
The robot used for the simulated experiments was the holonomic Robotnik SUMMIT XL equipped with a 2D LiDAR Hokuyo UST-10LX. 
It was commanded with stable linear and angular velocity of at approximately 1 m/s and 1 deg/s, respectively.

Using the \ac{URDF} model of this robot it is possible to leverage the different packages of the \ac{ROS} Navigation Stack for \ac{RViz}. 
One of these packages is NAVFN which assumes a circular robot and allows to plan a path from a start point to an endpoint in a grid based on a Costmap.

A Costmap is an inflated version of the given 2D \ac{OGM} with a specified amplification radius created to avoid the robot colliding with obstacles while navigating through the environment.

To speed up the usage of the \ac{OGM} for robot simulation, the Gazebo Plug-in PgmMap creator \shortcite{Yang.2018}, was also implemented, allowing the creation of maps with known origin position. 
In practice, this step is not required since the alignment between the real world and the map can be retrieved as a localization system result.

It is worth mentioning that using navigational goals instead of single movement commands is very convenient for data simulation since it significantly reduces the probability of collisions, which can make the entire sequence useless. 

Following this approach, 2D LiDAR, \ac{IMU}, Wheel odometry, and ground truth odometry were simulated in the six scenarios (three models with and without dynamic agents). 
The resulted trajectories of the simulation are presented in Figure \ref{fig:Sequences}.

\subsection{Implementation details}

Due to the stochastic nature of PF algorithms (AMCL and GMCL) and similarly as done by \shortcite{AlshikhKhalil.2021}, these methods were executed 30 times in each sequence, and the average values were calculated.

Similarly as \shortcite{Zimmerman.23.03.2022}, we consider that a method converges when its pose estimate is within a distance of \SI{0.5}{\metre} from the ground truth pose. If after the first \SI{95}{\percent} of the sequence, convergence does not happen, then it is considered a failure.

Unfortunately, SLAM Toolbox could not be evaluated for global localization since it does not provide this service. 
The lifelong mapping mode of SLAM Toolbox was also tested for the matter of completeness; however, it yields to unwanted results, with a poor performance.
\section{RESULTS AND ANALYSIS}
\label{chap:results}

\begin{table*}[ht]
  \centering
  \caption{Summary of the quantitative evaluation results for each sequence. Translational RMSE in centimeters and angular RMSE in degrees respectively.}
    \begin{tabular}{c|r|l|r|l||r|l|r|l||r|l|r|l}
    \toprule
    Method  & \multicolumn{2}{c|}{1-1} & \multicolumn{2}{c||}{1-2} & \multicolumn{2}{c|}{2-1} & \multicolumn{2}{c||}{2-2} & \multicolumn{2}{c|}{3-1} & \multicolumn{2}{c}{3-2} \\
    \midrule
    AMCL    & 8.49     & 0.44    & 8.47    & 0.50    & 33.68   & 2.71    & 37.44   & 3.26    & 63.04    & 3.29    & \multicolumn{1}{r|}{65.12} & 3.37 \\
    GMCL    & 8.27     & 0.24    & 7.86    & 0.24    & 24.27   & 2.57    & 52.38   & 4.37    & 66.60    & 3.70    & \multicolumn{1}{r|}{126.91} & 4.46 \\
    SLAM Toolbox & \textbf{3.69} & \textbf{0.17} & \textbf{3.95} & \textbf{0.17} & 28.69   & 1.50    & 23.57   & 1.50    & \textbf{37.84} & \textbf{1.34} & \multicolumn{1}{r|}{\textbf{37.96}} & \textbf{1.70} \\
    Cartographer & 4.01    & 0.24    & 3.96    & 0.25    & \textbf{7.19} & \textbf{0.15} & \textbf{4.11} & \textbf{0.21} & \textbf{-} & \textbf{-} & \textbf{-} & \textbf{-} \\
    \bottomrule
    \end{tabular}%
  \label{tab:results}%
\end{table*}%

The libraries provided by \shortciteN{Grupp.2017} and \shortciteN{Zhang.2018} were used to calculate the error metrics of the various methods on the different sequences.

\subsection{Pose tracking}

In Table \ref{tab:results} we present the translational and rotational \ac{RMSE} for each sequence for each method evaluated on the pose tracking problem with the ground truth from the simulation.

\begin{figure}[htb]
  \centering
        \includegraphics[width=0.5\textwidth]{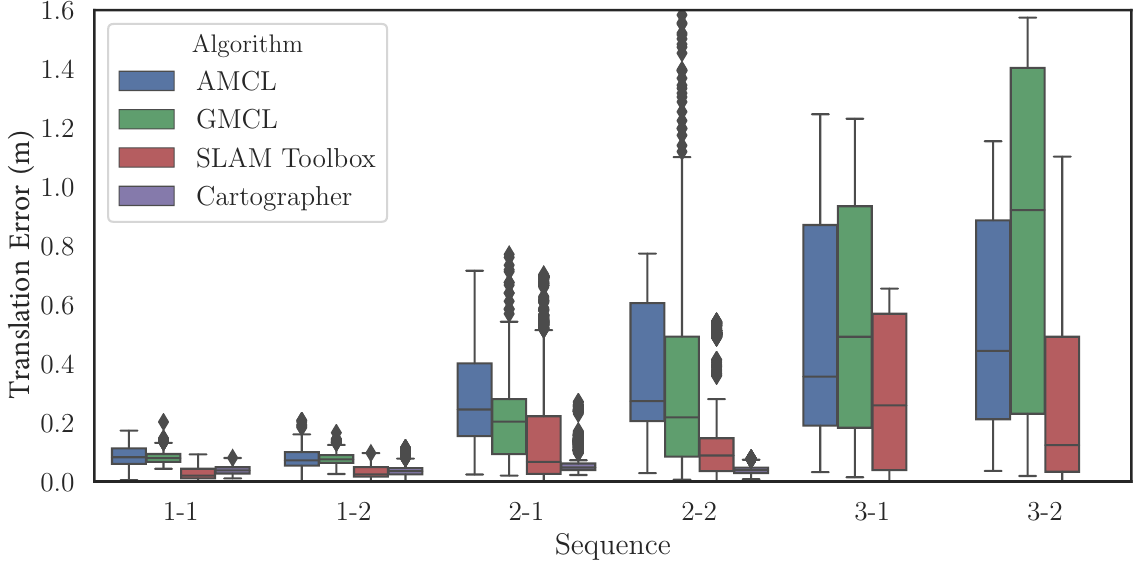}
\caption[Boxplot]{Statistics of the pose error estimates in translation for each method on the six evaluation scenarios.} \label{fig:BoxPlot_trans}
\end{figure}

Figure \ref{fig:BoxPlot_trans} presents a summary of the statistics of the translational errors for all the methods in all sequences.


Overall it can be seen that \ac{GBL} methods always perform better than \ac{PF} algorithms in the pose tracking problem.


Among the tested \ac{PF} algorithms, \ac{GMCL} performs most of the time better than \ac{AMCL}. 
Only in the scenarios 2-2, 3-1, and 3-2, \ac{AMCL} achieves lower RMSE.
In scenarios 2-2 and 3-2 \ac{GMCL} has a very high translational \ac{RMSE}. 

This shows that the additional filters of \ac{GMCL} cause the method to be more sensitive to dynamic environments in changing environments.


Regarding the \ac{GBL} algorithms, SLAM Toolbox achieves the best performance in scenarios 1 and 3. 
As expected, scenario 3 (with the most significant Scan-BIM deviations) was the most challenging scenario for all the methods.
On top of that, in this scenario, the pure localization mode of Cartographer always found wrong data associations, resulting in wrong relative constraints that cause localization failure. 
Therefore Cartographer could not be quantitatively evaluated in this environment, even when an initial approximated pose was provided.
Nonetheless, Cartographer achieved an impressive performance in the scenario 2 (real-world scenario), accomplishing a translational \ac{RMSE} four times lower than SLAM Toolbox,  in the environment without dynamic agents (\SI{7,19}{\centi\metre} and \SI{28,69}{\centi\metre} respectively) and almost six times lower in the scenario with dynamic agents (\SI{4,11}{\centi\metre} and \SI{23,57}{\centi\metre} respectively).

\nointerlineskip

\subsection{Global localization}

\begin{footnotesize}

\begin{figure}[h]
\centering{
\def\svgwidth{0.5\textwidth}
{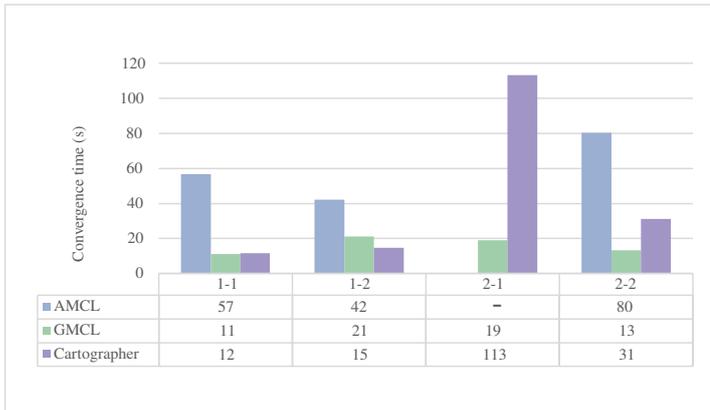}
\caption{Convergence time in seconds for the various methods in the different scenarios.}
\label{fig:convTime}
}
\end{figure}
\end{footnotesize}

The performance of the different methods regarding convergence time is presented in Figure \ref{fig:convTime}.

\ac{GMCL}, thanks to its Self-Adaptive \ac{PF}, performs the best in the global localization problem. 
Only in scenario 1-2 Cartographer shows a slight superiority. 
Meanwhile, \ac{AMCL} takes always at least twice as long compared to the other methods to converge to a good pose. In addition, it does not converge in scenario 2-1.

Due to the high level of Scan-BIM deviations, none of the implemented methods converges while trying to solve the global localization problem in scenario 3. 
\section{CONCLUSIONS}
\label{chap:conclusions}

In this paper, besides contributing with methods to create \ac{OGM}s from BIM models and transforming them to pose graph-based maps for robust localization, we provide an extensive comparison of diverse state-of-the-art localization 2D-LiDAR algorithms in three different levels of Scan-BIM deviations, with and without dynamic agents.

We found that \ac{GBL} algorithms overperform \ac{PF} algorithms in the pose tracking problem. 
In the case of a map with very low (or negligible) Scan-BIM deviations, SLAM Toolbox achieves the best performance. 
On the contrary, if the map has a medium level of Scan-BIM deviations (for example, due to large pieces of furniture or as-planned and as-built differences), as in a real-world office building, Cartographer is the best performing method.
However, in a case where the level of changes in the environment is too high (such as in scenario 3), SLAM Toolbox, while with a relatively high error, would be the best option among the tested localization algorithms.

The fact that \ac{PF} algorithms only consider the most recent observation to update the belief of the current pose gives them certain robustness to deal with high ambiguity scenarios (such as scenario 3). 
However, it also causes high inaccuracies when the level of Scan-BIM deviations is medium (such as in scenario 2).
On the other hand, \ac{GBL} algorithms taking advantage of a recent history of observations, can handle better this real-world scenario and can track the robot's pose more accurately.

Nonetheless, \ac{GMCL} perform better for the global localization problem than \ac{GBL} algorithms.

In general we recommend using a \ac{GBL} algorithm for accurate BIM-based (or floor plan-based) 2D LiDAR pose tracking in real-world environments and \ac{GMCL} for global localization.

To facilitate the correct implementation of \acl{GBL} algorithms, we contribute with an open-source method to create efficiently accurate pose graph-based maps from any \ac{OGM}. 

In addition, we provide with a method to create \ac{OGM}s from complex multi-story BIM models, which additionally can be leveraged for path planning and autonomous navigation. 

State-of-the-art SLAM techniques have switched from using particle filters to graph-based optimization approaches, based on our experiments, we can conclude that it will be analogously advantageous for most localization systems.

\section{FUTURE WORK}

In the light of the experimental results and motivated by related research, we believe that the following are promising future research directions:

Consider not only 2D-LiDAR information but also 3D-LiDAR sensor data is a promising direction to reliably handle large levels Scan-BIM deviations, as partially shown by \shortciteN{Blum.2021} and \shortciteN{Moura2021}.

Fuse multiple sensor modalities, such as \ac{IMU}, RGB-D cameras, together with LiDAR sensors, would increase the robustness of a localization method to deal with fast angular movements and deprecated scenarios, as demonstrated by \shortciteN{Lin.11.09.2021} and \shortciteN{Xu.2022}. 
    
To achieve major robustness, the extraction of detailed information from a BIM model, such as the position of room numbers labels, doors and windows can support to solve the global localization problem even in symmetric environments, as done by \shortciteN{Zimmerman.23.03.2022} and \shortciteN{Haque.2020}.

\section{ACKNOWLEGEMENTS}
The presented research was conducted in the frame of the project ``Intelligent Toolkit for Reconnaissance and assessmEnt in Perilous Incidents'' (INTREPID) funded by the EU's research and innovation funding programme Horizon 2020 under Grant agreement ID: 883345.

\bibliographystyle{00_Template/chicaco}
\bibliography{03_References/references.bib}

\end{document}